\documentclass[journal]{IEEEtran}

\usepackage[colorlinks,urlcolor=blue,linkcolor=blue,citecolor=blue]{hyperref}

\usepackage{color,array}
\usepackage[noadjust]{cite}
\usepackage{subcaption}
\usepackage{graphicx} % graphics
\usepackage[capitalise]{cleveref}
\usepackage{stfloats}

\title{Explaining Genetic Programming Trees using Large Language Models

}

\author{
    Paula~Maddigan,
    Andrew~Lensen,~\IEEEmembership{Member,~IEEE,}
	Bing~Xue,~\IEEEmembership{Fellow,~IEEE}%
	\thanks{This work was supported by the University Research Fund at Te Herenga Waka--Victoria University of Wellington under grant number 410128/4223.}%
	\thanks{The authors are with the Centre for Data Science and Artificial Intelligence, and School of Engineering and Computer Science;  Victoria University of Wellington; Wellington 6140; New Zealand (e-mail: paula.maddigan@ecs.vuw.ac.nz; andrew.lensen@ecs.vuw.ac.nz; bing.xue@ecs.vuw.ac.nz;).}%  
}%

\begin{document}
\maketitle

\begin{abstract}
Genetic programming (GP) has the potential to generate explainable results, especially when used for dimensionality reduction. In this research, we investigate the potential of leveraging eXplainable AI (XAI) and large language models (LLMs) like ChatGPT to improve the interpretability of GP-based non-linear dimensionality reduction. Our study introduces a novel XAI dashboard named GP4NLDR, the first approach to combine state-of-the-art GP with an LLM-powered chatbot to provide comprehensive, user-centred explanations. We showcase the system's ability to provide intuitive and insightful narratives on high-dimensional data reduction processes through case studies. Our study highlights the importance of prompt engineering in eliciting accurate and pertinent responses from LLMs. We also address important considerations around data privacy, hallucinatory outputs, and the rapid advancements in generative AI. Our findings demonstrate its potential in advancing the explainability of GP algorithms. This opens the door for future research into explaining GP models with LLMs.
\end{abstract}


\makeatletter
\setlength{\@fptop}{0pt}
\setlength{\@fpbot}{0pt plus 1fil}
\makeatother

\section{Introduction}
\label{introduction}
\IEEEPARstart
{G}{enetic} programming (GP) is a powerful evolutionary computation technique that evolves computer programs to solve complex tasks. Its versatility and ability to automatically discover model structure make it an attractive choice for solving many problems. GP is capable of producing functional mathematical mappings with good predictive accuracy. These symbolic mappings (trees) are a promising approach for enabling eXplainable Artificial Intelligence (XAI) \cite{Mei2023}. 

The field of XAI is at the forefront of current research. It is crucial within sectors such as medical diagnosis and financial risk assessments, where explainability is required to gain trust among stakeholders \cite{Quinn2022, Salahuddin2022, Kuiper2022}. However, even with the symbolic nature of GP, understanding the semantics of a GP model/tree or the meaning of individual features may require expert domain knowledge. Even understanding the functionality of the evolutionary process may lie beyond the comprehension of end-users. 

The term \textit{end-users} is deliberately vague. Different audiences need wildly different explanations, personalised to their background and requirements. Ribera \cite{ribera2019}  highlighted the importance of approaching XAI from a \textit{user-centred} perspective. They categorised the targeted audience into three broad groups: \textit{developers and AI researchers}, \textit{domain experts}, and \textit{lay users}.  They showed that explanations are multifaceted, requiring different explanations for every user group. For example, vocabulary needs to be adapted to match the comprehension level of each group, by omitting technical terms for lay users and integrating domain-specific terminology when engaging with experts. Humans are also social creatures, who learn through conversation \cite{Miller2019}. An explanation delivered through a \textit{conversational exchange} would allow users to directly request answers suited to their own domain knowledge and technical background, greatly improving the explanation quality.

The proliferation of large language models (LLMs) such as OpenAI's ChatGPT has powered a notable surge in chatbot development, facilitating conversational question-and-answering over a broad range of domains.  Therefore, this study introduces an AI-driven chatbot to explain the functionality of GP models/trees. Leveraging LLMs in this way capitalises on a wealth of domain knowledge, aiding in understanding results. When responses do not align with the user's level of understanding, they may seek further clarification through conversation. The inherent nature of the LLMs enables users from diverse backgrounds to pose questions around presented findings by using language, vocabulary, and grammar of their preference. Existing studies highlight the multilingual capabilities of LLMs \cite{maddigan2023b} and their comprehension in understanding questions containing grammatical or typographical errors \cite{maddigan2023a}. 

The versatility of genetic programming deems it applicable to a plethora of tasks in real-world applications, including but not limited to symbolic regression \cite{Haider2024}, job scheduling \cite{Nguyen2024}, classification \cite{Fan2024}, and feature selection \cite{Ain2024}.  This paper focuses specifically on improving the explainability of Genetic Programming for Nonlinear Dimensionality Reduction (GP-NLDR) methods. Modern datasets often have thousands or tens of thousands of features, which can only be processed by extremely complex and expensive machine learning approaches \cite{wu2022, Nguyenqv2022, Fernstad2020, Agrawal2022, Lensen2021b}. NLDR methods can greatly reduce the dimensionality (number of features) of a dataset, making the data easier to process and understand. GP-NLDR, unlike traditional NLDR methods, has shown promise in performing \textit{explainable} NLDR, where the reduced dimensions (embedding) can be directly understood in the context of the original features \cite{Lensen2021b, Uriot2022, Lensen2019, Lensen2020}. In this paradigm, each new dimension in the embedding is represented by a single GP tree, where the tree takes a subset of original features as its inputs (leaves) and produces a single output (embedding dimension). Despite continued research, GP-NLDR can still produce overly complex trees, which are not explainable to non-experts. 

This study proposes GP4NLDR, a web-based dashboard that utilises an LLM-powered chatbot to explain GP-NLDR trees. We opt for a web-based architecture to enhance the  accessibility of our research to the diverse audience  identified in our study. Leveraging an intuitive graphical user interface with rich visualisations simplifies interaction with the system, contrasting with alternative delivery methods such as command-line processes and code libraries.  While we constrain our study's scope to GP-NLDR, our framework is applicable to many GP applications, laying the groundwork for significant advances in explainable GP.

\subsection*{Major Contributions}
\begin{itemize} 

\item This study explores the feasibility of using LLMs such as ChatGPT to provide human-like explainability of GP expressions. It contributes to combining the fields of evolutionary computation and generative AI, a notably scarce approach in existing literature. 
We demonstrate that our proposed methodology can be extended to other applications of GP. 

\item Previous work \cite{Lensen2019, Lensen2020, Lensen2021} has presented state-of-the-art techniques for GP-NLDR. This study makes this research accessible by making our custom-built online %application 
system GP4NLDR\footnote{https://gp4nldr.streamlit.app/} publicly available. The platform allows users to learn about GP-NLDR by running it on datasets using different fitness functions and run parameters. The GP expressions and trees are viewable together with run results.

\item Our proposed approach incorporates LLM-driven conversational interactions via a chatbot natural language interface. The chatbot is customised through prompt engineering and retrieval augmented generation to help strengthen the understanding of tree expressions and output. The GP4NLDR software interface allows the use of the chatbot with self-generated examples or through pre-loaded case studies.  

\item Finally, we contribute to the growing body of research highlighting limitations in using LLMs and the impact of hallucinations on XAI, with a unique perspective on these issues within explainable GP.
\end{itemize}

\section{Related Work on XAI}
\label{sec:relatedwork}

Recent years have seen the emergence of diverse XAI techniques fostering the explainability of black-box models. Comprehensive analyses \cite{Longo2024,Ali2023} present the complexities and nuances of these numerous XAI strategies across broad interdisciplinary domains. Our focus is not to re-visit the extensive list already presented by the authors, but rather highlight some as illustrative examples supporting the goal of our research. In predictive machine learning models, approaches such as SHAP \cite{Lundberg2017} and LIME \cite{Ribeiro_2016} provide insights for local and global explainability; Anchors \cite{alibi2021} provides a set of rules under which predictions still hold with confidence; and DiCE \cite{dice2020} is used in modelling \textit{what-if} counterfactuals. Previous studies demonstrate their use in domains such as healthcare \cite{Venturini2024,Maddigan2022,Hussain2024} and education \cite{Afrin2023,Raji2024}. However, these approaches target model developers capable of translating the interpretations into lay terms for communicating to stakeholders. Prior studies \cite{kuzba2020,guimaraes2022,nguyen2022} have developed chatbots for end-users to engage in conversational exchanges, enhancing their understanding of these XAI tools' output.  However, no studies have utilised groundbreaking large language models such as ChatGPT within this domain.  

There is extensive literature that seeks to improve the explainability of GP \cite{Mei2023} through approaches such as building smaller trees with bloat control \cite{Luke2006} or using fewer features \cite{Tran2016}.  However, this poses the same challenges with XAI tools previously discussed, where the output is targeted towards those knowledgeable in these concepts, failing to enable XAI from a \textit{user-centred} perspective which caters to a broader, non-expert audience.  

Communicating the explainability of AI systems has also been explored from a social sciences standpoint. Previous studies \cite{Miller2019} highlight how the field of XAI may benefit from incorporating insights from philosophy, cognitive psychology/science, and social psychology to understand how humans define, generate, and evaluate explanations. Their work highlights how XAI may benefit from understanding how decisions are explained to humans and how humans articulate decisions to each other. 

The role of natural language in generating explanations has been surveyed in prior studies \cite{Cambria2023}. The authors concluded only a handful of recent XAI approaches either considered natural language explanations for end-users or implemented a method capable of generating such explanations. A recent review of works in the emerging field of interpreting LLMs and using them for explanation highlights LLMs possess the opportunity to redefine interpretability across a wide range of applications \cite{Singh2024}. A recent study \cite{Sartori2024} proposes leveraging large language models for the automated analysis of optimisation algorithms within a web-based tool \cite{ChaconSartori2023} for the generation of search trajectory networks . They highlight how this application of LLMs may enhance the user experience of the tool and bridge the knowledge gap for those without prior understanding of the application.  However, no previous work has been identified using natural language chatbots to delve deeper into explaining GP expressions, nor its use in the field of NLDR. Several notable context-based chatbot implementations have recently emerged in other domains, leveraging similar technologies to those implemented in our study. Aisha \cite{Lapp2023}, a library chatbot, uses prompt engineering with a Chroma vector database together with LangChain and ChatGPT to deliver reference and support services to students and faculty through a Streamlit interface.  In the medical domain, accGPT \cite{Rau2023} is a ChatGPT-based chatbot that provides personalised imaging recommendations supporting clinical decision-making.  It leverages LlamaIndex to access information within the American College of Radiology documentation.

\section{Methodology}
\label{sec:methodology}
In this study, we used Streamlit\footnote{https://streamlit.io/}, an open-source Python framework, to build an online web-based application GP4NLDR\footnote{Hosted on Streamlit Community Cloud https://gp4nldr.streamlit.app/}. The application incorporates existing GP-NLDR code bases from prior works in the field \cite{Lensen2019, Lensen2020, Lensen2021} to perform the NLDR. The process outputs one GP tree for each dimension of the new embedding, together with performance metrics. We then introduce the use of generative pre-trained transformer (GPT) LLMs to facilitate conversational question answering \cite{Zaib2022}, to greatly improve the explainability of the trees found by GP-NLDR. We further developed our approach by incorporating intelligent prompt engineering and pre-initialising the LLM with additional knowledge from existing literature through the use of retrieval augmented generation, which guides it to deliver focused and targeted on-topic responses. We utilise the popular Langchain \cite{langchain} framework to streamline the integration of LLMs and the workflow components.  

\begin{figure*}[tbp]
\vspace{-4em}
\centering
\includegraphics[width=\textwidth]%[width=2.5in]
{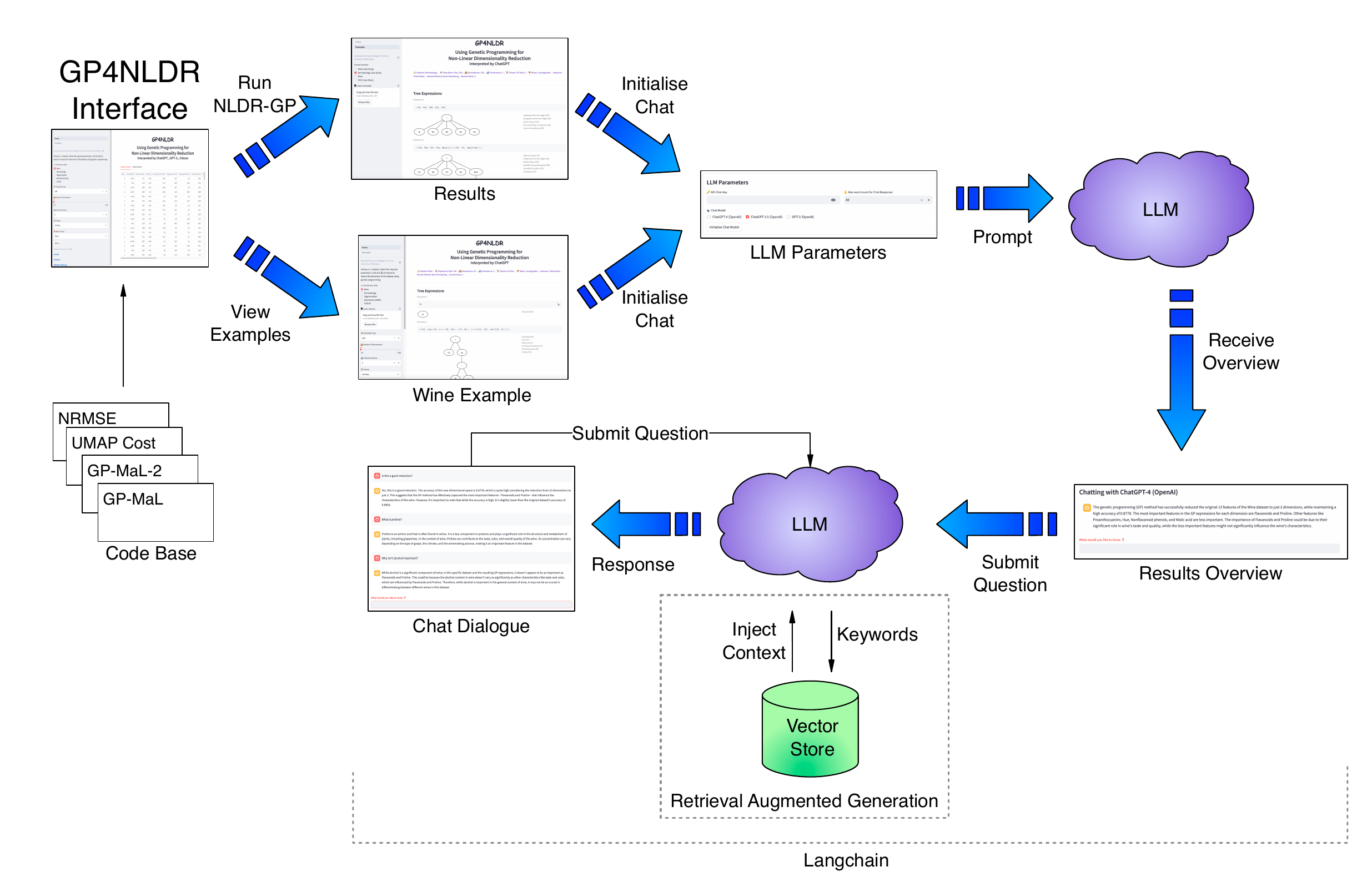}
\caption{Overview of GP4NLDR Architecture }
\vspace{-1em}
\label{overview}
\end{figure*}

\cref{overview} depicts an overview of the GP4NLDR architecture.  The 
system provides a facility to run the NLDR-GP process on a given dataset or view pre-loaded examples for quick use. After results are generated, the chat feature can be initialised. A written summary of the process is presented as interpreted by the LLM. Then, further dialogue conversations with the chatbot can commence.  We elaborate on these stages more comprehensively in the following subsections.

\subsection{GP4NLDR System}

\begin{figure*}[tbp]
\vspace{-1em}
\centering
\includegraphics[width=\textwidth]{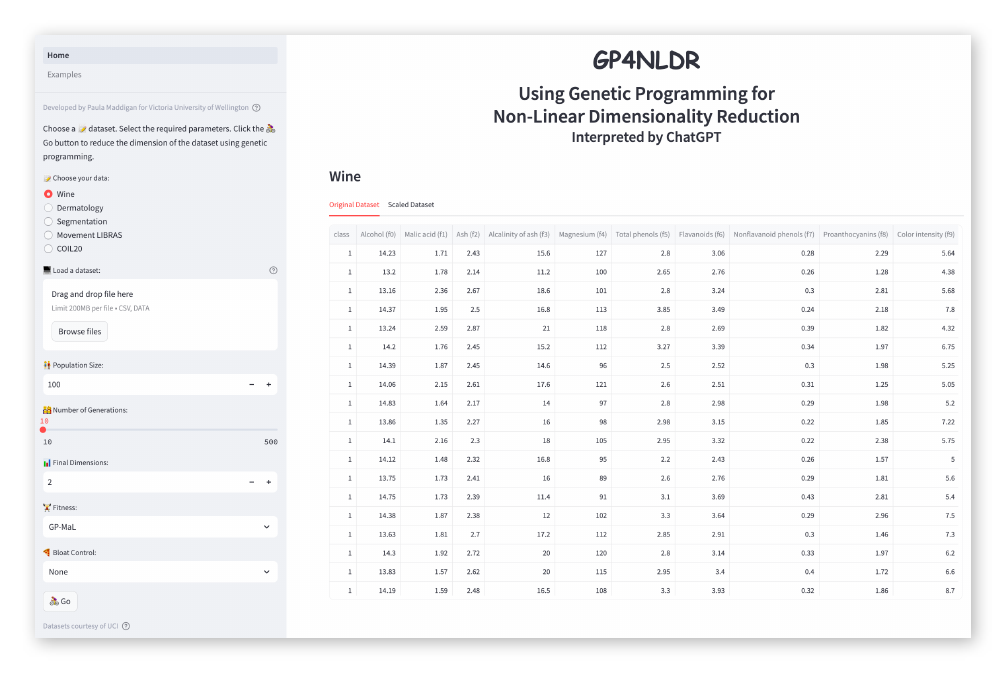}
\caption{GP4NLDR System}
\vspace{-1em}
\label{main_page}
\end{figure*}

\cref{main_page} depicts the GP4NLDR system showing parameter options and dataset information. For ease of understanding the dataset, the original values are presented, along with the scaled data\footnote{Data scaled using Scikit-learn's \texttt{MinMaxScaler}.} used in the  dimensionality reduction process.  We now discuss the design of each part of the system in turn.

\subsubsection{Parameter Options for the NLDR process}
\begin{itemize}
    \item Population Size: the number of individuals in the population. A larger size may enhance the learning ability but increases computational complexity. A smaller size may lead to insufficient diversity and premature convergence.
    \item Number of Generations: how many iterations of the algorithm to perform. It requires a balance between allowing the population to evolve towards an optimal solution and avoiding extended computational costs. Monitoring convergence on the fitness plot may help determine a suitable value.
    \item Final Dimensions: how many dimensions the embedding should contain (i.e.\ the number of GP trees). Prior knowledge of the data domain or task requirements determines this number. Alternatively, for visualisation of the dataset, three or fewer dimensions would be chosen.  
    \item Fitness: The fitness function measures the quality of the NLDR solution and helps guide the evolutionary process towards a better solution. Available options include GP-MaL \cite{Lensen2019}, GP-Mal-2 (the first objective of \cite{Lensen2020}), UMAP Cost \cite{Lensen2021} and NRMSE \cite{Lensen2021}.
    \item Bloat Control: optional techniques to help reduce the size of GP trees to prevent unnecessary growth, improving performance and tree interpretability. Options include: (1) lexicographic \cite{luke2002lexicographic} --- a parsimony pressure method that prefers smaller trees when fitness values are equal; (2) double tournament \cite{luke2002fighting} --- uses two tournaments: one for fitness and one for size, with the selection of which tournament is run first and the probability that a smaller individual is chosen over a larger more complex one; and (3) Tarpeian \cite{poli2003simple} --- which penalises large individuals during evolution according to a provided probability.
\end{itemize}

\subsubsection{Display of NLDR Results}
On completion of the GP-NLDR process, the results are displayed 
%on the \texttt{Home} page 
for analysis.  A summary of parameters is noted, followed by tree expressions and visualisations for each new dimension. The raw embedding result is presented alongside a plot depicting fitness per generation. If the embedding dimensionality is 3D or lower, a visualisation of the embedding is provided: either as a 3-D rotational plot, a 2-D scatter plot, or a 1-D bar graph. A random forest classifier \cite{Breiman2001} implemented in Scikit-Learn \cite{scikit-learn} with 10-fold cross-validation is also used to provide an estimated accuracy for both the original dataset and that of the new embedding, as a proxy of embedding quality.  

\subsubsection{Chatbot}
The chat feature is initialised upon entering a valid OpenAI API key, selecting an LLM (e.g. GPT-3.5 or GPT-4) for conversation, and confirming the approximate word limit for responses. The word limit is set to a default of 80 words. Too few words may return insufficient explanations. Excess words may prolong response times and introduce verbosity, repetition, or tangential answers.  

The pre-engineered prompt (discussed further in \cref{promptEngineering}) and initial question \textit{"Provide an exciting summary of the results"} are submitted to the LLM. The LLM returns a brief overview of the results as a starting answer for the conversation. Then, two-way conversational question-and-answering with memory retention begins, utilising retrieval augmented generation when required. At any stage, the results and chat history may be downloaded, allowing for reloading at a later point in time.

\subsubsection{Pre-loaded Examples}
The system provides exploration of previously generated GP-NLDR evolutionary runs, including the case studies presented in this work. The chat feature is available within each example to help further interpret the output.  This facility allows for the reproducibility of our research for each GP-NLDR case study presented. LLMs are deterministic models, fundamentally generating the same outputs for the same inputs. Nevertheless, as their responses are probabilistic, they may appear non-deterministic.  Therefore, it may not be feasible to achieve identical explanations even though the input prompt remains unchanged. If desired, previously generated results from user experiments may also be reloaded here for further analysis.

\subsection{Large Language Models}
The rapid advancement in LLMs throughout this research project opened avenues to investigate the capabilities of both existing and emerging models, including open-source solutions. Following the evaluation of the performance and accessibility for the task at hand\footnote{The evaluation process lies outside the contribution of this work and as such is not presented.}, OpenAI's ChatGPT-3.5 model (gpt-3.5-turbo) was adopted as a foundation for the development of GP4NLDR's chat feature. 
This \textit{Chat Generative Pre-trained Transformer} model 3.5 is based on the transformer deep learning architecture. It is designed to generate human-like text in response to input questions. This state-of-the-art language model excels at natural language processing and conversational exchanges. 

We used the Python \texttt{openai} library to facilitate an authenticated connection to the OpenAI models with requests submitted via the API endpoint. We refrained from explicitly including the model version suffix\footnote{For example gpt-3.5-turbo-0613} allowing us to take advantage of the continuous model upgrades, therefore insuring we provided the safest and most capable model version. OpenAI regularly upgrade model versions, thus for the long-term viability of our research it was important to mitigate deprecation issues stemming from tying the research to specific model versions. 
Additional options are provided within the chatbot for using the legacy model GPT-3 and the most recent addition, GPT-4. We use the default LLM model parameter settings, with the temperature set to zero to encourage response consistency.  To access the models in the chat function, a valid OpenAI API key is required\footnote{OpenAI API key available at https://openai.com/}.

\subsection{Retrieval Augmented Generation}
At the time of writing, ChatGPT-3.5 was trained on data up until the end of September 2021.  Consequently, with some research information beyond its reach or in publicly unavailable studies, many recent concepts in the evolving GP field are unknown to the model. Retrieval augmented generation (RAG) \cite{Lewis2020} is a technique to address this limitation. RAG builds a vector store/database of vector embeddings from relevant documents. By performing vector searching using similarity metrics, relevant information is extracted and injected as contextual background information into the user's prompt. This helps fill knowledge gaps in the model and provide it with recent insights, and presents a cost-effective and dynamic alternative to pre-training or fine-tuning models.
 
The GP4NLDR processor centres on the articles referenced in previous studies \cite{Lensen2019, Lensen2020, Lensen2021}, but can be easily extended to other methods. A vector store of these papers was constructed by generating vector embeddings of the documents, and then made available to the application. A computationally expensive vector database\footnote{Vector databases provide create, read, update, and delete functionality.} was not needed for this use case, and so we opted to use FAISS \cite{faiss}, Facebook's AI Similarity Search vector index library\footnote{https://faiss.ai/}. Given a fixed number of stored articles, with no requirement to add additional files or update existing ones, FAISS is a very efficient and suitable option. 
The vector store is integrated into the application chat feature for OpenAI models.  During conversational chatting, user-provided questions are analysed against a pre-defined set of keywords: \texttt{gp-mal, gpmal, gpmal2, gp-mal2, gp-mal-2, tarp, lexi, tourn, umap, nrmse}\footnote{The keyword list is further customisable in the configuration settings of the application software.}. In our initial prototypes (without a vector store), using these keywords often returned responses of limited usefulness, even on occasion provoking hallucinations, as these abbreviations are less prevalent within the model training data. When questioned about these keywords in the context of GP through the ChatGPT OpenAI interface, the LLM did not consistently provide accurate responses. Hence, should these keywords be present, RAG is activated, and the FAISS vector index is queried to fetch relevant background information. Upon retrieval, the information is injected into the prompt. For queries outside the keyword list, it is expected that the model maintains enough background knowledge and the prompt is sufficient to acquire an informative response to the query.  This process can be seen within \cref{overview}.

\subsection{Prompt Engineering}
\label{promptEngineering}

\begin{figure}[tbp]
\vspace{-1em}
\centering
\includegraphics[width=3.4in]
%[width=3in]%[width=2.5in]
{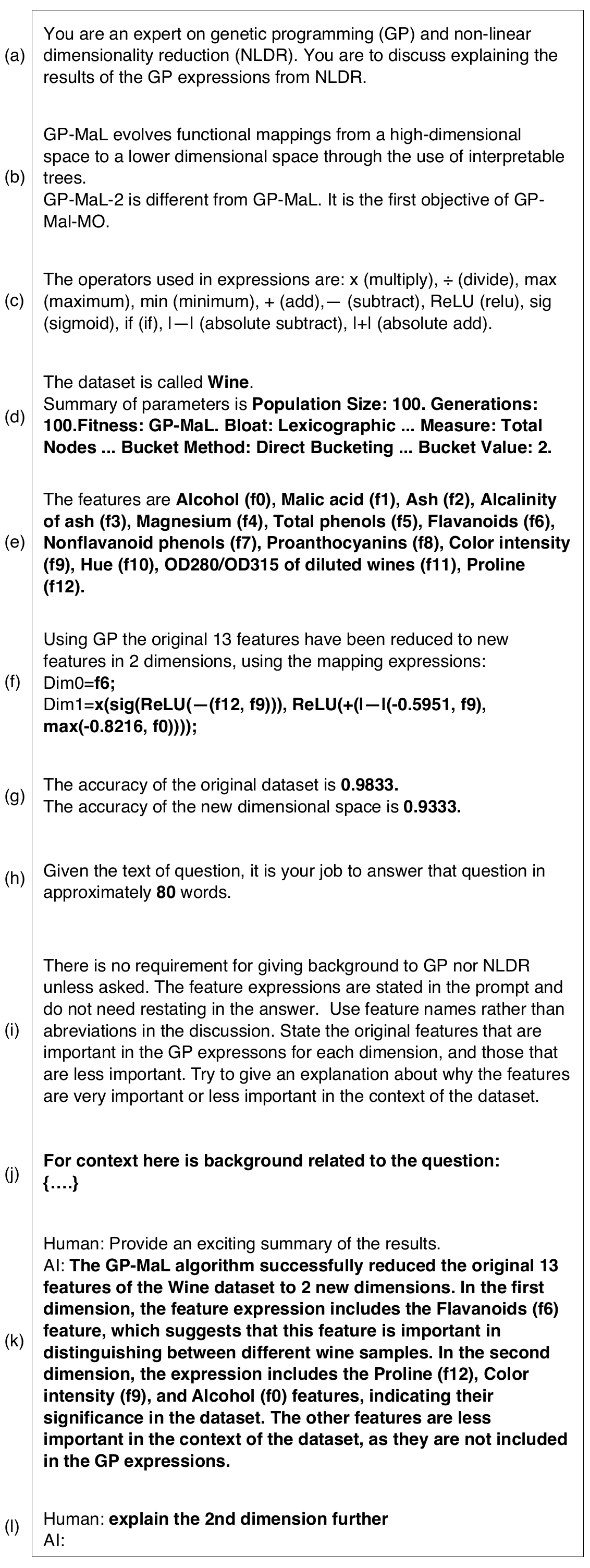}
\caption{Prompt Example}
%\vspace{-1em}
\label{prompt}
\end{figure}

Careful consideration was given to our prompt development to elicit informative and consistently reliable responses. \cref{prompt} shows the entire initial prompt using an example from the Wine Case Study presented later in our results. Bolded text represents the automatic injection of content from the specific example.

\begin{itemize}
\item \cref{prompt}(a) 
establishes the context for the discussion, directing the LLM to focus on genetic programming and non-linear dimensionality reduction.  
\item The fitness function GP-MaL-2  is not explicitly mentioned within the publications in the vector store. Consequently, we define it explicitly in the prompt. 
\item \cref{prompt}(c) explains the operators used in the GP algorithm. 
\item \cref{prompt}(d) informs the LLM with the name of the dataset and a summary of the parameters used.
\item The dataset features are listed in \cref{prompt}(e). Should the feature list exceed 40,  we replace the feature list with the text \textbf{f0 to f{n-1}} (as a dataset with \textbf{n} features). This tweak avoids exceeding the token limit for large datasets, such as COIL20, with more than 1000 features. 
\item \cref{prompt}(f) provides the LLM with the dataset dimension specifications and resulting expressions.
\item Providing the classification accuracy of the original and reduced space in \cref{prompt}(g) informs the LLM how well the NLDR process performed.
\item Specifying the response word count when initialising the LLM allows flexibility in token usage during chatting, with the allowance specified within the prompt in \cref{prompt}(h).
\item \cref{prompt}(i) guides the LLM further in the expectations for response content, ensuring that information in the prompt is not repeated.
\item Should the question contain keywords, background information is retrieved from the vector store and injected in \cref{prompt}(j).
\item \cref{prompt}(k) requests an initial response from the LLM to provide an overview of the results.
\item \cref{prompt}(l) shows an example initial conversational chat dialogue between the \texttt{Human} and the \texttt{AI}.
\end{itemize}

\subsection{LangChain}
LangChain \cite{langchain} provides a modular framework for building applications powered by LLMs. The toolkit offers flexibility for integrating a diverse range of LLM model variants.
Its versatile structure and functionality facilitated the integration of RAG into our application workflow. Preserving conversational memory within chatbots is paramount, and LangChain seamlessly facilitated the memory retention process. 

\subsection{GP4NLDR Evaluation}
\label{eval_section}
We demonstrate the capabilities of GP4NLDR and chat interactions over three case studies. The first two case studies are based on the Wine \cite{wine} and Dermatology \cite{dermatology} datasets, which contain meaningful feature names. The final case study uses the larger COIL-20 \cite{coil} dataset with 1024 features, lacking descriptive feature names. These examples demonstrate the system's behaviour across a range of different parameter options:
\begin{itemize}
    \item A small dataset with 13 features and 178 instances through to a larger dataset of 1024 features and 1,440 instances.
    \item Different fitness functions (GP-MaL and GP-MaL-2).
    \item Reducing to two or three final dimensions. 
    \item The use of lexicographic bloat control compared to no bloat control.
    \item From 100 generations through to 1000 generations.
\end{itemize}
Furthermore, these case studies investigate the calibre of chatbot responses in the following situations:
\begin{itemize}
    \item Supplying descriptive feature names compared to sequentially allocating non-descriptive feature names, which may limit background information.
    \item Using keywords such as gpmal to engage with RAG.
    \item Subjective questioning, for instance, querying how good results are.
    \item Asking questions using terminology not identical but similar to feature names in the dataset.
    \item Probing the importance of features.
    \item The multilingual capabilities of LLMs.
\end{itemize}

In presenting the evaluation of each case study, we showcase subsections of the system results for illustration while depicting the complete interface in the Appendix. 
It is not feasible to demonstrate all possible parameter settings and scenarios.  A curated selection has been chosen, emphasising those deemed most meaningful in showcasing our research results.  

We pose questions in a manner that aligns with lay users. This demographic of users stand to benefit most from our study as they more typically rely on an intermediary party to translate existing ML explainability tools into summary text. The results are evaluated manually by comparing the correctness of the generated chatbot responses to the results depicted in the GP expression trees. In addition, we manually evaluate and discuss the quality and accuracy of the responses to queries unrelated to the trees, which more specifically target  dataset domain, GP, or NLDR questions. As the work presented in this study is the first to use LLMs to provide explanations of GP trees, difficultly lies in benchmarking our approach and providing measurable metrics of accuracy. 

Our developed system\footnote{https://gp4nldr.streamlit.app/} has been made publicly available for further experimentation and testing.  The presented case study examples are viewable within the application, and may be further analysed using the chatbot. However we note, as touched on earlier, the generation of identically worded responses from subsequent questioning using the same prompt may not be achievable due to the inherent nature of LLMs. In this work we perceive this as an advantage, imparting a sense of personalised responses to the user rather than generic explanations. 

\section{Results}
\label{sec:results}
\subsection{Wine Case Study}
\label{subsec:wine}
\begin{figure}[tbp]
\centering
\vspace{-2em}
\includegraphics
%[width=\textwidth]
[width=3.4in]
%[width=2.5in]
{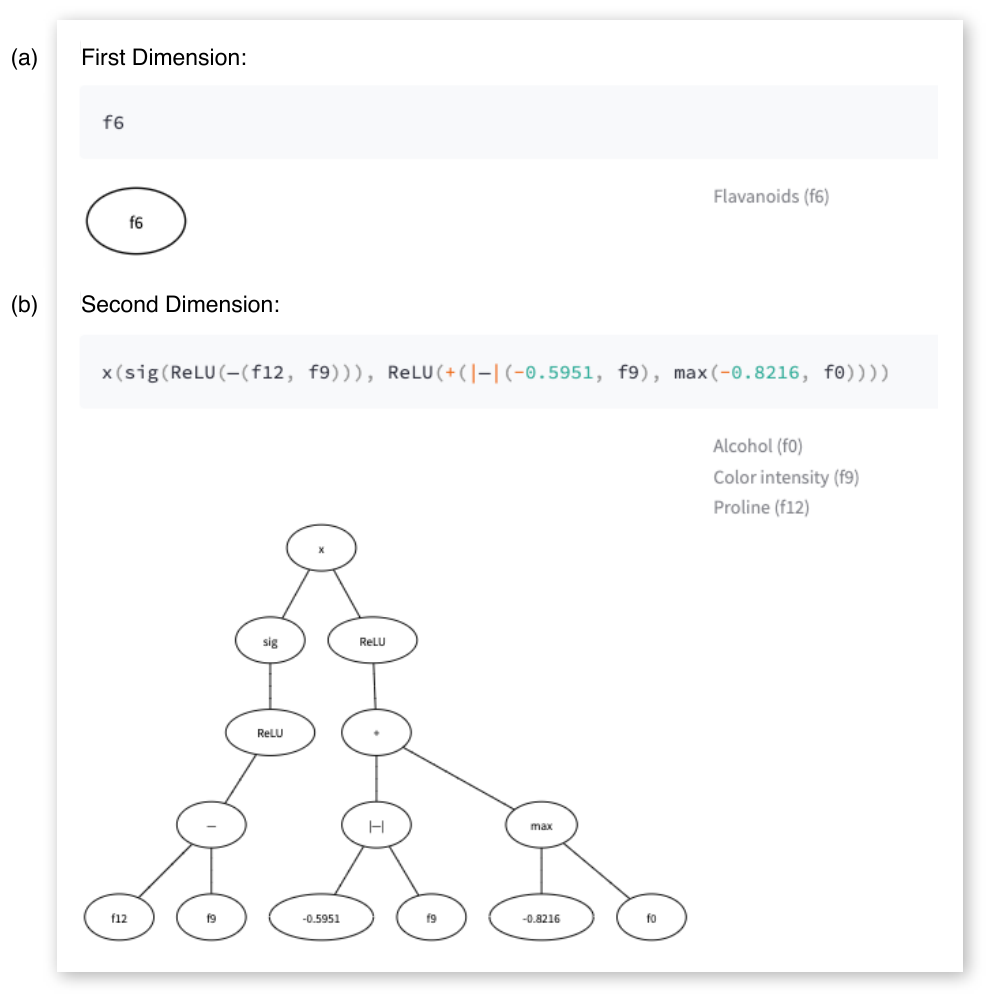}
\caption{Wine Case Study Trees}
\vspace{-1em}
\label{CaseStudy1_Trees}
\end{figure}
\begin{figure}[tbp]
\centering
\includegraphics
%[width=\textwidth]
[width=3.3in]
%[width=2.5in]
{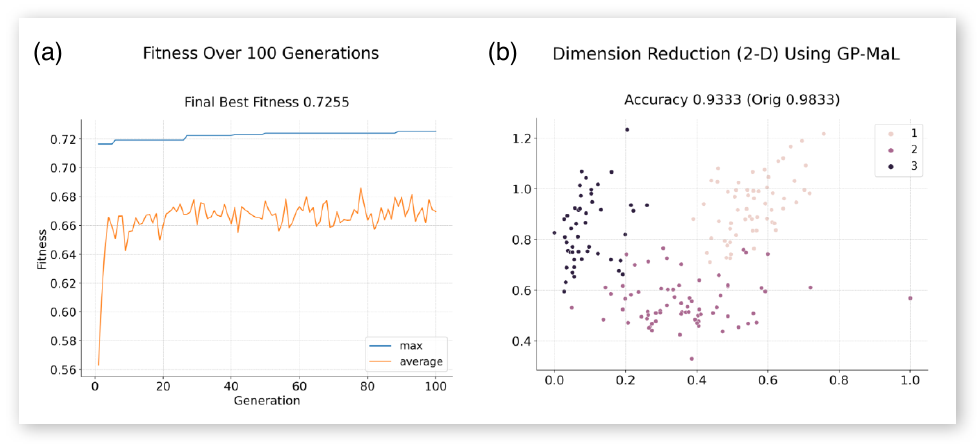}
\caption{Wine Case Study Plots}
\vspace{-1em}
\label{CaseStudy1_Plots}
\end{figure}
 The Wine dataset\footnote{https://archive.ics.uci.edu/dataset/109/wine} consists of 13 features and 178 instances detailing the chemical analysis of three types of Italian wine. 
 The GP4NLDR process is run using the GP-MaL fitness function, reducing the Wine dataset to two dimensions after 100 generations using a population size of 100 and lexicographic bloat control.  The first embedding dimension (GP tree) shown in \cref{CaseStudy1_Trees}(a) is a single node, representing the flavonoids feature. The second tree utilises the Proline, colour intensity, and alcohol features in \cref{CaseStudy1_Trees}(b). The fitness plot in \cref{CaseStudy1_Plots} shows the function converges quickly, and the three classes are easily distinguishable within the 2D embedding plot. %

 \begin{figure}[tbp]
 \vspace{-2em}
\centering
\includegraphics
%[width=\textwidth]
[width=3.4in]
%[width=2.5in]
{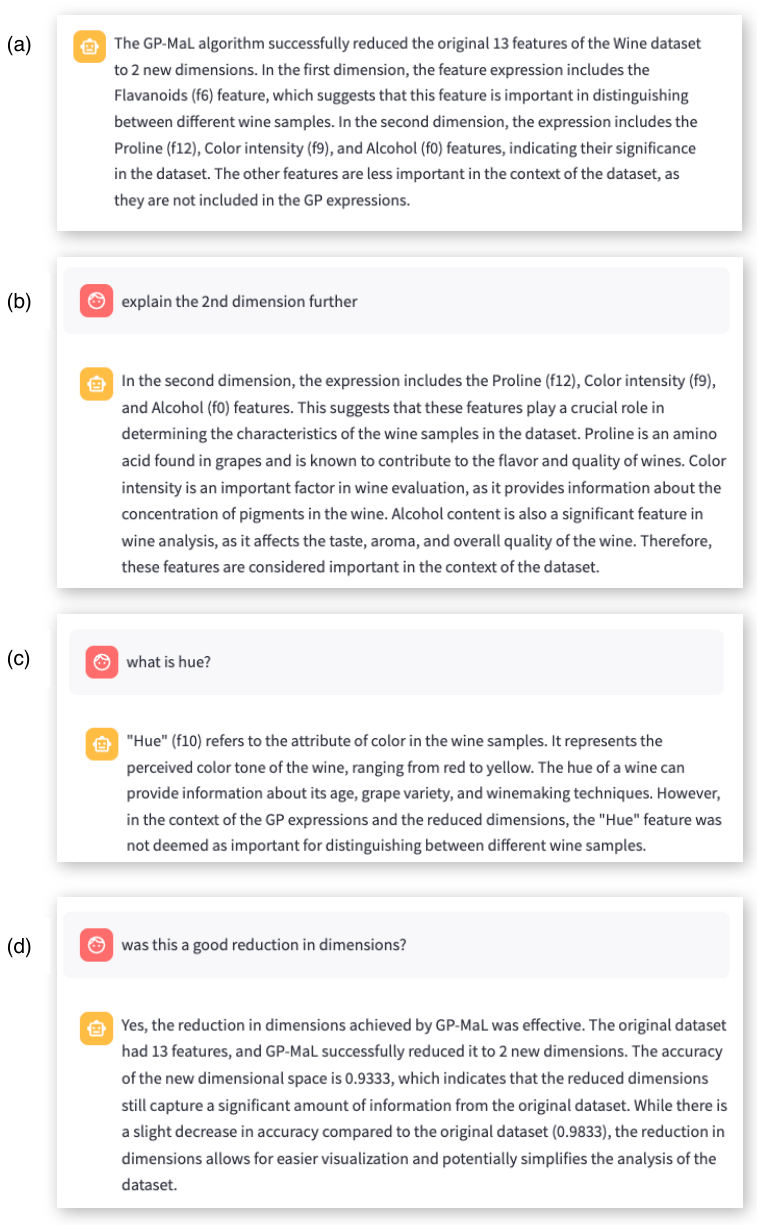}
\caption{Wine Case Study Chat}
\vspace{-1em}
\label{CaseStudy1_Chategs}
\end{figure}

 Initiating the chatbot allows further investigation into the results.
 Including the feature names within the dataset structure feeds the LLM additional context when considering why some features are present in the new dimensional space while others remain absent.  The overview initially generated is illustrated in \cref{CaseStudy1_Chategs}(a). A brief discussion of the dimensionality reduction is given together with noting the features present in each dimension and deemed important. In \cref{CaseStudy1_Chategs}(b), when asked to \textit{``explain the 2nd dimension further"}, the LLM expands the justification for the inclusion of each feature by providing definitions of the features and their relationship to the dataset. When questioning the LLM about a specific feature \textit{``what is hue?"} in \cref{CaseStudy1_Chategs}(c), the LLM gives an overview of its definition followed by its contribution to the results. In this example, hue was not part of the embedding and hence not deemed as important.  
 
 Supplying the accuracy of the embeddings when classified by a random forest algorithm may give somewhat subjective opinions from the LLM when asked if it is a \textit{``good"} reduction. \cref{CaseStudy1_Chategs}(d) shows the LLM believes this example is \textit{``effective"} with \textit{"a slight decrease in accuracy"} from 0.9833 to 0.9333.  

\subsection{Dermatology Case Study}
The Dermatology dataset\footnote{https://archive.ics.uci.edu/dataset/33/dermatology} with 34 features classifies the type of erythematous-squamous disease into six classes (psoriasis, seborrheic dermatitis, lichen planus, pityriasis rosea, chronic dermatitis, and pityriasis rubra pilaris.) 12 features are clinical evaluations with a further 22 histopathological features from skin samples. There are 358 instances in total. 
\begin{figure}[tbp]
\centering
%\vspace{-1em}
\includegraphics
%[width=\textwidth]
[width=2.5in]
%[width=2.5in]
{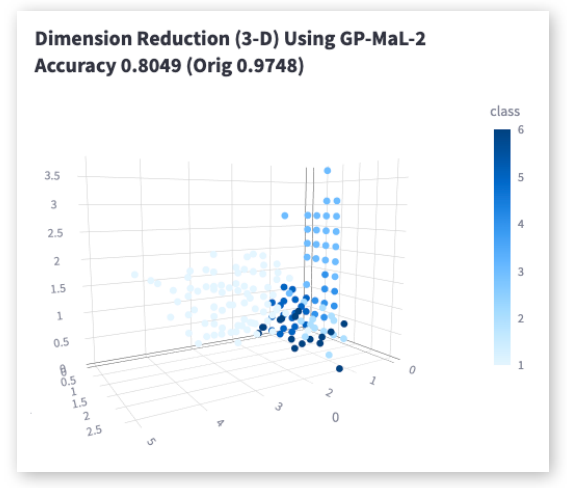}
\caption{Dermatology Case Study Plot}
\vspace{-2em}
\label{CaseStudy2_Plots}
\end{figure}

The GP4NLDR process is run using the GP-MaL-2 fitness function, reducing the Dermatology dataset to 3 dimensions over 200 generations using a population size of 100 and lexicographic bloat control. \cref{CaseStudy2_Plots} illustrates the 3-D plot representing the reduced embedding. The three new dimensions each use between four and six features.

\begin{figure}[tbp]
\vspace{-1em}
\centering
\includegraphics
%[width=\textwidth]
[width=3.4in]
%[width=2.5in]
{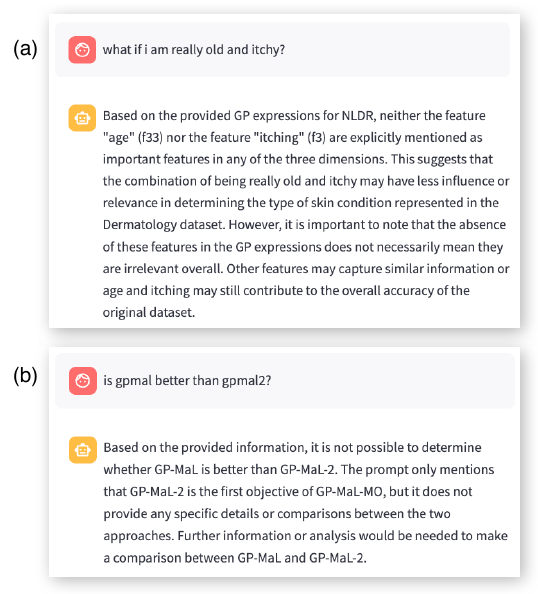}
\caption{Dermatology Case Study Chat}
\label{CaseStudy2_Chategs}
\vspace{-1em}
\end{figure}

\begin{figure}[tbp]
\centering
\includegraphics
%[width=\textwidth]
[width=3.4in]
%[width=2.5in]
{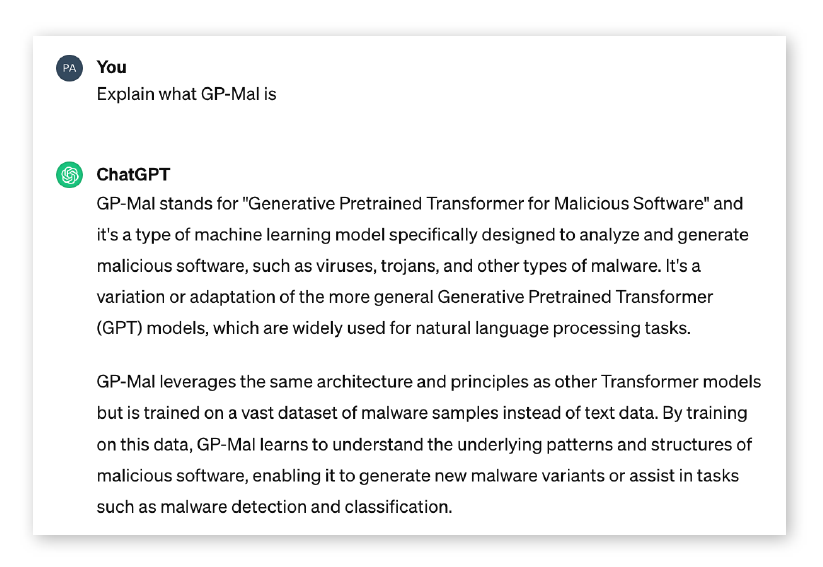}
\caption{ChatGPT Web Interface https://chat.openai.com/}
\label{chatgpt_interface}
\vspace{-1em}
\end{figure}

Once more, including the feature names within the dataset structure has assisted the chatbot in providing contextual conversational exchanges.  When asked \textit{``What if I am really old and itchy?"}, the LLM maps the word \textit{``old"} to the feature \texttt{age} and the word \textit{``itchy"} to the feature \texttt{itching}. \cref{CaseStudy2_Chategs}(a) shows it subsequently responds that neither of these features appears in the tree expressions and hence they have less influence or relevance in determining the type of skin condition.

In asking \textit{``Is GP-MaL better than GP-MaL-2?"} in \cref{CaseStudy2_Chategs}(b), the LLM notes it has not been supplied with information detailing specific comparisons between the two functions.  This response may be perceived as uninformative; nonetheless, it shows that clever prompt engineering can help deter hallucinations.  Throughout prompt development, experiments demonstrated the LLM's susceptibility to generating inaccurate information. \cref{chatgpt_interface} illustrates this concern by asking the question ``\textit{Explain what GP-Mal is}" through the online ChatGPT Web Interface\footnote{https://chat.openai.com/}. ChatGPT 
 invents the definition ``\textit{Generative Pretrained Transformer for Malicious Software}" and endeavours to discuss it convincingly. This example is testament to the mitigation of such hallucinatory outcomes by incorporating tailored prompt engineering.

\subsection{COIL-20 Case Study}
The Columbia Object Image Library (COIL-20)  dataset\footnote{https://www.cs.columbia.edu/CAVE/software/softlib/coil-20.php} consists of gray-scale images of 20 objects. For each object, a photographed image was captured every 5 degrees as it was rotated 360 degrees on a motorised turntable, giving 72 images for each object. Each of these images represents one row in the dataset, giving 1,440 rows in total for the 20 objects. The original 128x128 pixel image size is cropped to 32x32 pixels, producing a 1024-dimensional feature example. The absence of feature names necessitates generically assigned attribute labels f0 to f1023. The full results in the supplementary material illustrate the results from the NLDR process, producing somewhat large trees. The fitness plot depicts the gradual improvement in the function over the 1000 generations. In reducing the dimensions from 1024 to 2, accuracy has decreased from 0.9868 to 0.6375.

\begin{figure}[tbp]
\vspace{-3em}
\centering
\includegraphics
%[width=\textwidth]
[width=3.4in]
%[width=2.5in]
{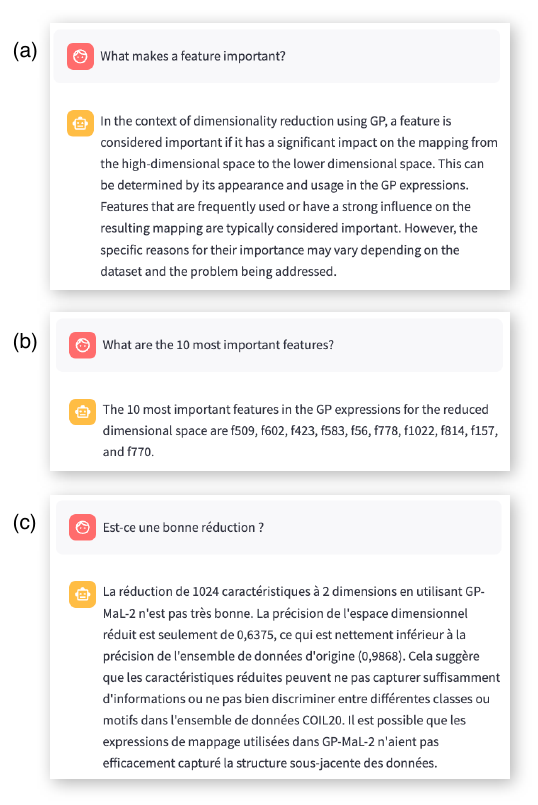}
\caption{COIL Case Study Chat}
\vspace{-1em}
\label{CaseStudy3_Chategs}
\end{figure}

Using the chatbot, we ask \textit{``What makes a feature important?"}. In \cref{CaseStudy3_Chategs}(a), the LLM informs us important features have a significant impact on the mapping from the high-dimensional space to the low-dimensional space. It continues to explain the importance is determined by the feature's appearance and usage in the expressions. Frequently used features or those having a strong influence on the mapping are considered important. 

To further explore the importance of features in the dataset, we ask the LLM in \cref{CaseStudy3_Chategs}(b) to name the ten most significant features. Initially, it assesses the first dimension. f509 appears twice, with one instance high up in the tree, having a significant influence on the outcome. f602 also appears high up in the tree, presents itself 4 times, and is on both sides of the root \texttt{max} node. f423 is featured in the next level, and f583 appears five times, with two instances in the subsequent level. Moving to the second dimension, f56 with 6 occurrences has 2 of these instances high up in the tree on both sides of the root \texttt{max} node. f778 with 4 occurrences sits alongside f56 on the 3rd level of the tree. f1022, f814, f157, and f770 complete the top 10.

To highlight the multilingual capabilities of the chatbot, we ask it in French \textit{``Est-ce une bonne réduction?"} meaning \textit{``Is this a good reduction?"}. The chatbot also responds in French, as shown in \cref{CaseStudy3_Chategs}(c).
Using Google Translate, we can translate this back to English, which gives us:   
\textit{``Reducing 1024 features to 2 dimensions using GP-MaL-2 is not very good. The precision of the reduced dimensional space is only 0.6375, which is significantly lower than the precision of the original dataset (0.9868). This suggests that reduced features may not capture enough information or discriminate well between different classes or patterns in the COIL20 dataset. It is possible that the mapping expressions used in GP-MaL-2 did not effectively capture the underlying data structure"}. The multilingual ability of LLMs is a significant opportunity for making advances in AI accessible to a wider audience.
\section{Discussion}
\label{sec:discussion}

The experiments in this study confirm the effectiveness of an XAI dashboard in communicating the results of GP-NLDR. Leveraging LLMs such as ChatGPT effectively contributes to user-centred explanations through conversational chatbot technology.  Employing AI-powered web-based applications such as GP4NLDR draws on the latest cutting-edge research delivering state-of-the-art tools to individuals. In this section, we discuss several of our findings in more detail. We believe that aspects of this discussion could be very useful in guiding the development of methods integrating GP and LLMs.

\textbf{Prompt engineering} is a dynamic and evolving field requiring careful crafting to steer models towards relevant and accurate responses. Recently, it has gained significant attention due to its pivotal role in shaping the behaviour of LLMs. The trend towards formalising prompt structures has given rise to defining prompt techniques such as zero-shot, few-short, chain-of-thought, tree-of-thought, and more. In this work, we adopt a combination of techniques. Structured prompting can effectively maintain a uniform tone in chatbot responses. However, in our setting, this is not of paramount concern. Our developed prompt, although slightly verbose and unstructured, introduces novelty and diversity, enabling the LLM to craft its responses creatively if desired. Avoiding explicitly requesting a fixed response structure, such as bullet points, sentences, paragraphs, or abbreviations, contributes to enhancing engagement with the chatbot. Furthermore, in targeting a \textit{user-centred} approach, we do not seek to impose restrictions on response style, which may potentially hinder ingenuity and interest when generating explanations.  However, potentially allowing the user an option to indicate their level of comprehension may facilitate a more tailored response tone, which could be addressed in future work. 

\textbf{Data privacy} concerns within LLM-powered applications continue to be at the forefront of discussions in research and industry. End-users interacting with AI applications should seek reassurance in knowing the confidentiality and security of their data is maintained, especially sensitive and personal information.  In this work we demonstrated within the prompt template no raw data is transmitted to the LLM, only the dataset name and feature list.  Nonetheless, this does not prevent the user from entering sensitive information and transmitting it voluntarily.

\textbf{Hallucinations} are a growing concern in developing applications integrating LLMs. Throughout the development of this work, we witnessed entirely fictional information returned from the models following questioning. 
To reduce the incidence of hallucinations, we integrated our tailored prompt template with retrieval augmented generation (RAG). This generally addressed this problem, but, unfortunately, no robust solution has yet been identified to circumvent these situations. Throughout the development of the prompt, we encountered guardrails imparting superfluous advice not pertinent to providing further explanations. Ongoing research in LLMs is expected to address this. 
We also note other recent concerns arising with the use of LLMs in applications such as adversarial attacks \cite{Zhou2024} and bias within the models and their benchmarks \cite{McIntosh2024}.  It is not within the scope of this work to delve into these issues further, but we acknowledge these challenges are of ongoing concern and necessitate further research.

\textbf{Rapid advancements} are frequently seen in the fast-moving domain of generative AI. During the preparation of this paper, recent announcements such as ChatGPT Enterprise have been regularly released. The edition boasts an extended token limit of 32k (4 times the current capacity), enterprise-grade privacy and security, and the expansion of model knowledge through integration with company data\footnote{https://openai.com/blog/introducing-chatgpt-enterprise}. Within our research setting, an increased token limit may enable the complete list of feature names for higher-dimensional datasets to be included in the LLM prompt, eliminating the need for truncation. Occasionally, exceptionally lengthy GP expressions may surpass the current token limit; an extended token limit may be advantageous in some scenarios.  With the addition of extra privacy measures, the prompt could be enriched to include a subset of dataset rows. Supplementing the model with this information may enhance the conversational explanations.  The facility to integrate company data would be an ideal alternative to using RAG.  These innovations will continue to address ongoing concerns in developing AI-driven applications.  

\textbf{Future work} could delve deeper into the use of LLMs for explaining GP expressions in other fundamental machine learning tasks. Exploring other retrieval methods and/or alternative vector store approaches has the potential to further improve the efficacy of our framework. In addition, exploring alternative architectures to the Langchain framework used in this study may offer further avenues for harnessing LLMs. Our work touched on the feasibility of leveraging other open-source LLMs. Further development of tailored prompt templates and consideration of fine-tuning these models could be advantageous in assessing their performance comparison with ChatGPT. Exploring the extension of chat parameters, such as explicitly targeting different audiences, may improve user experience and the understanding of explanations.  To more rigorously validate our research, future work will include human evaluation of the explanations \cite{vanderlee2019}.  Through user-group experiences we may assess the quality of results on a larger scale and endeavour to provide measurable benchmarks for use in subsequent research within this domain.

\section{Conclusion}
 \label{sec:conclusion}
This study presented a novel dashboard application to explain the results of GP-based nonlinear dimensionality reduction. Our proposed approach cohesively incorporates a variety of techniques, including a user interface, visualisation, a large language model chatbot, retrieval augmented generation, and prompt engineering to provide a system that greatly improves the explainability of GP. This is the first study of its kind encapsulating these elements within a unified system, spanning the domains of evolutionary computation and generative AI. We presented three robust case studies to highlight the usability of
our research in this field.  Incorporating a chatbot built on groundbreaking LLM techniques provides significant improvements to the explainability of GP expressions, with potential implications for the wider GP community. Furthermore, we have highlighted how leveraging LLMs for conversation provides a \textit{user-centred} approach accommodating the needs of a diverse audience. Our work has contributed to the gap in research around leveraging generative AI in explainable evolutionary computation.

\bibliographystyle{IEEEtran}

\bibliography{refs}
\appendices 
\begin{figure*}[!t]
\begin{center}
{\fontsize{18}{20}\selectfont Appendix}\\
%\Huge Appendix %%HAHAHAHAHA %%Thought you'd like that
\end{center}
\end{figure*}
%\section
\begin{figure*}[!h]
%\vspace{-5em}
\centering
\includegraphics[width=.9\textwidth]%[width=3in]%[width=2.5in]
{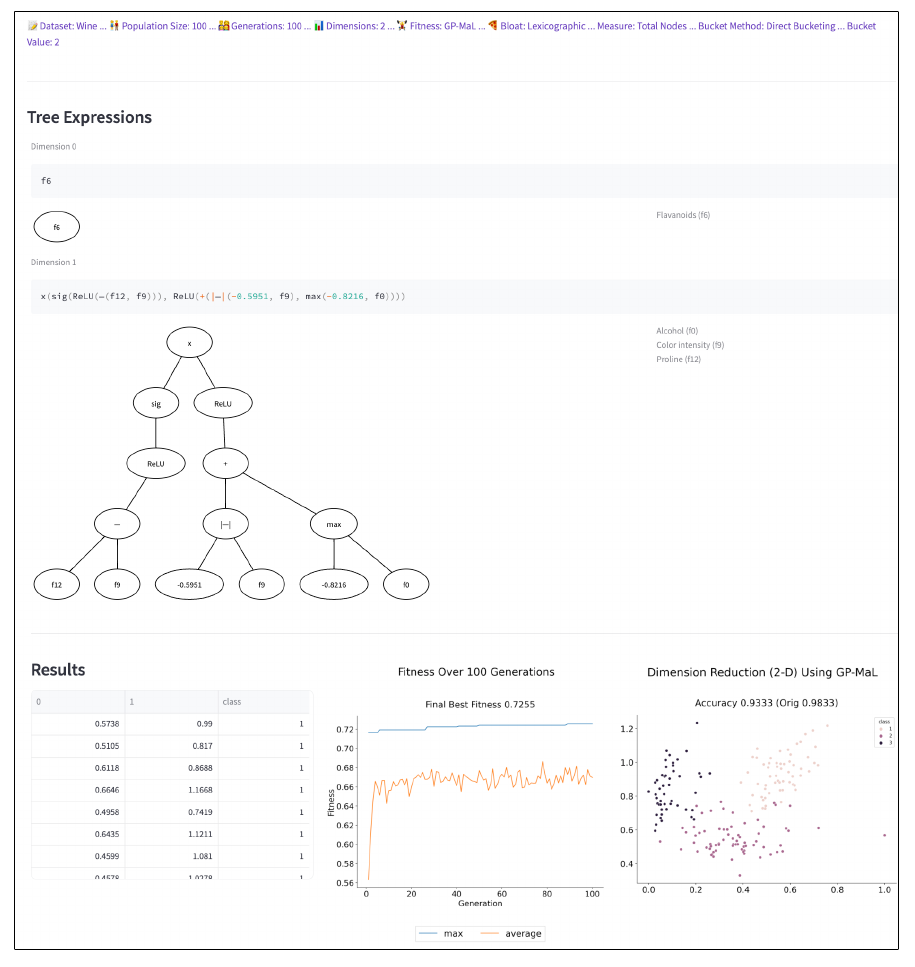}
\caption{ Wine Case Study Results from GP4NLDR}
\label{CaseStudy1_Results}
%\vspace{-15em}
\end{figure*}
\pagebreak
\begin{figure*}[!ht]
\centering
\includegraphics[width=\textwidth]%[width=3in]%[width=2.5in]
{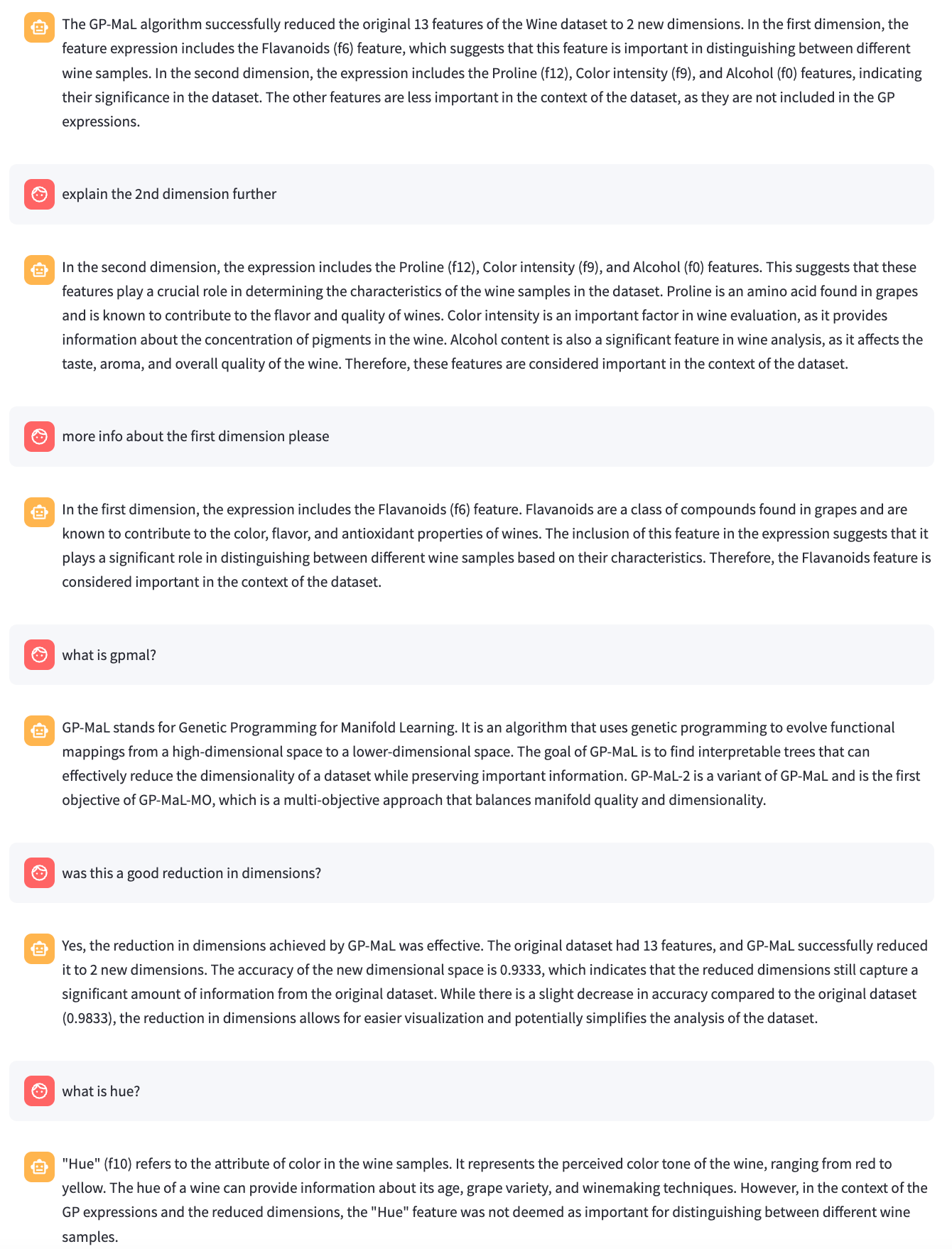}
\caption{Wine Case Study Chat Conversation Examples}
\label{CaseStudy1_Chat}
\end{figure*}

%\clearpage
%\section{Dermatology Case Study Output}
 \begin{figure*}[!ht]
\centering
\includegraphics%[width=\textwidth]
[width=\textwidth]
%[width=2.5in]
{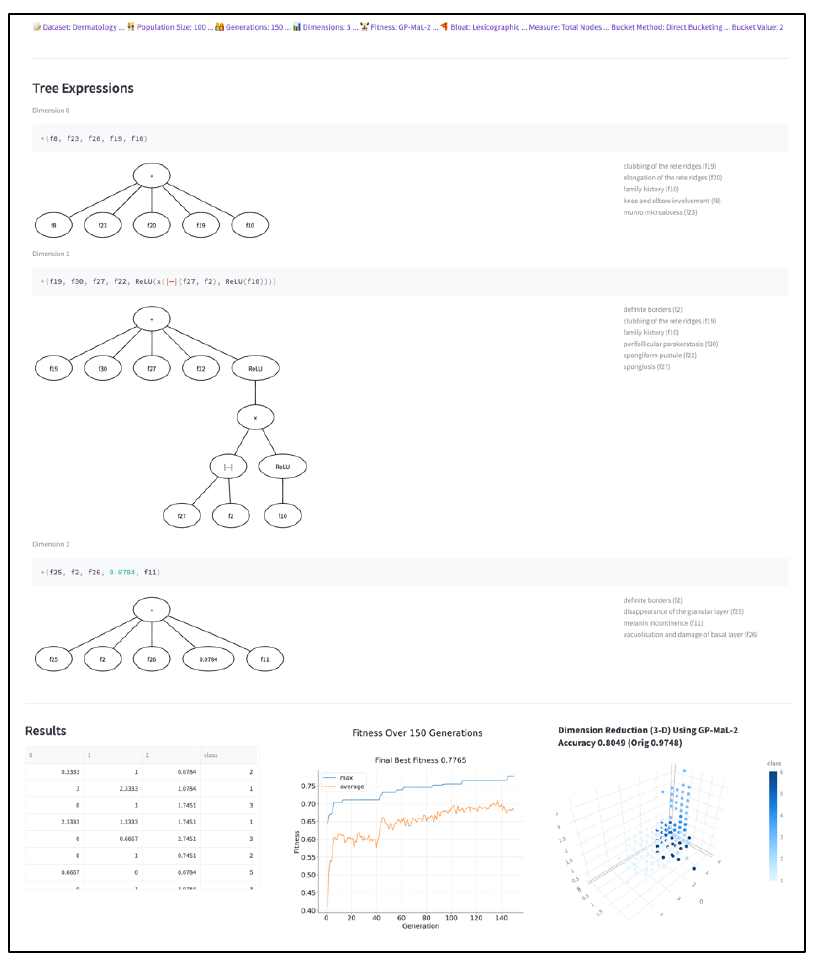}
\caption{ Dermatology Case Study Results from GP4NLDR}
\label{CaseStudy2_Results}
\end{figure*}

\begin{figure*}[!ht]
\centering
\includegraphics[width=\textwidth]%[width=3in]%[width=2.5in]
{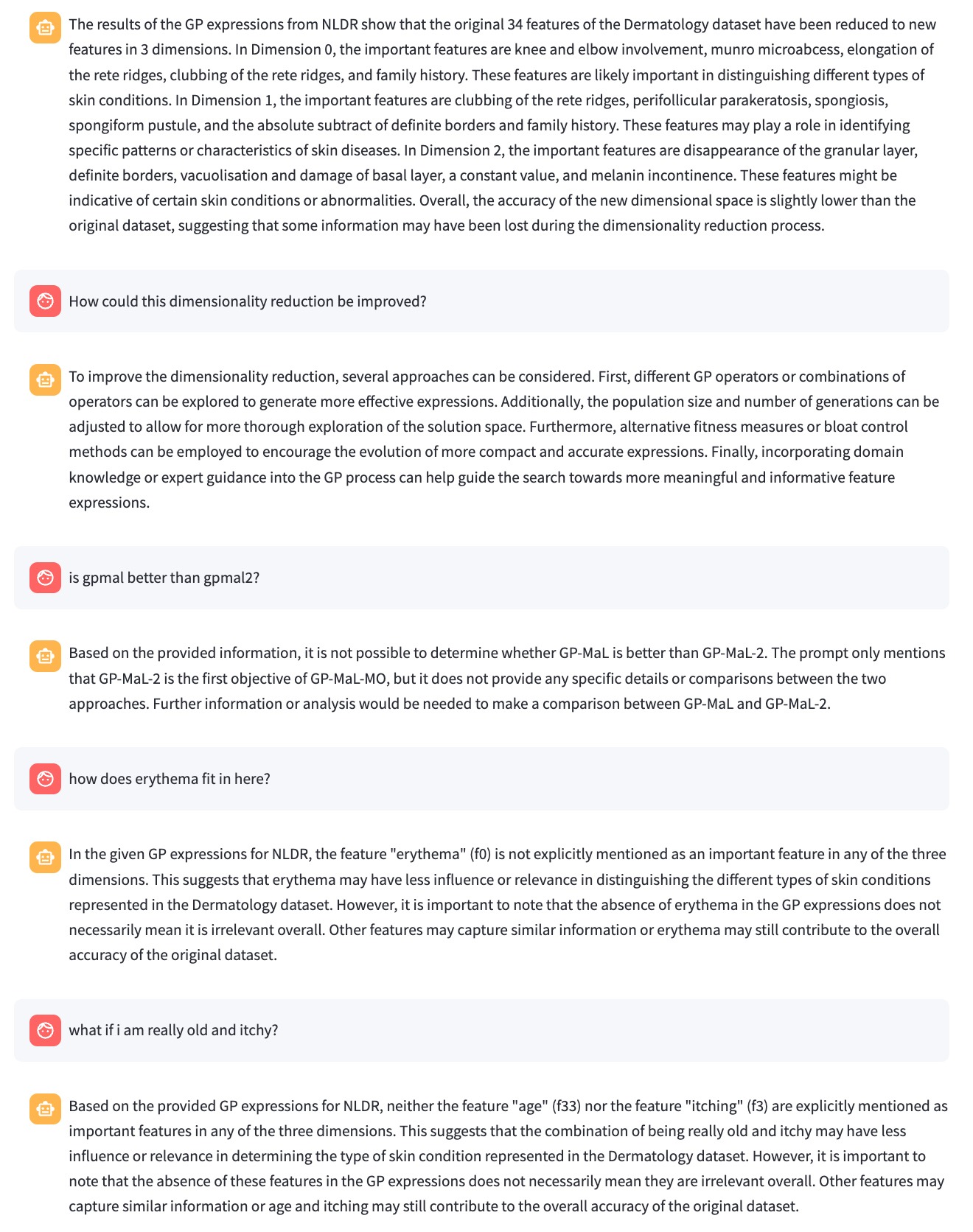}
\caption{Dermatology Case Study Chat Conversation Examples}
\label{CaseStudy2_Chat}
\end{figure*}

%\clearpage
%\section{COIL20 Case Study Output}
 \begin{figure*}[!ht]
\centering
\includegraphics[width=5in]
%[width=2.5in]
{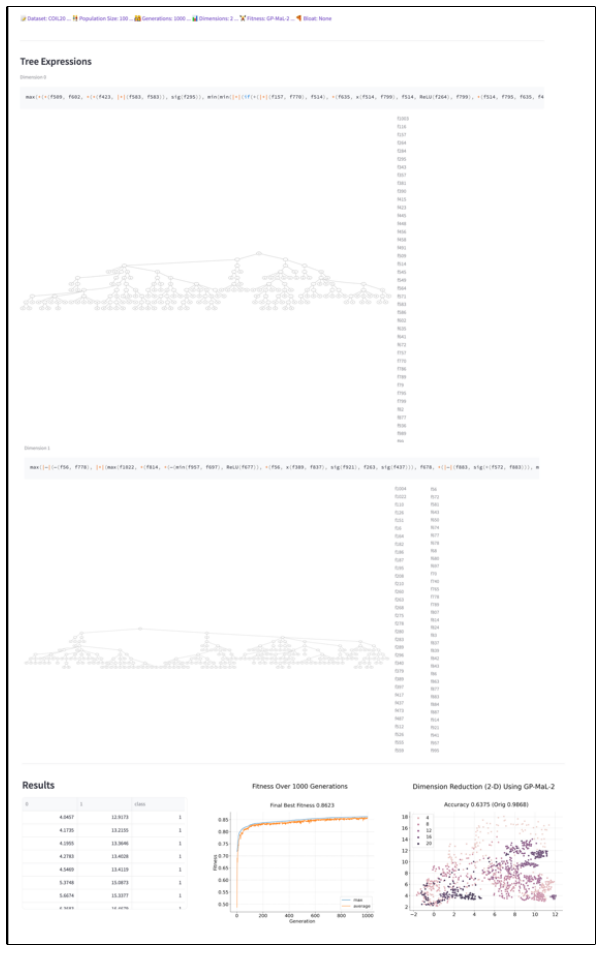}
\caption{ COIL20 Case Study Results from GP4NLDR}
\label{CaseStudy3_Results}
\end{figure*}

\begin{figure*}[!ht]
\centering
\includegraphics[width=\textwidth]%[width=3in]%[width=2.5in]
{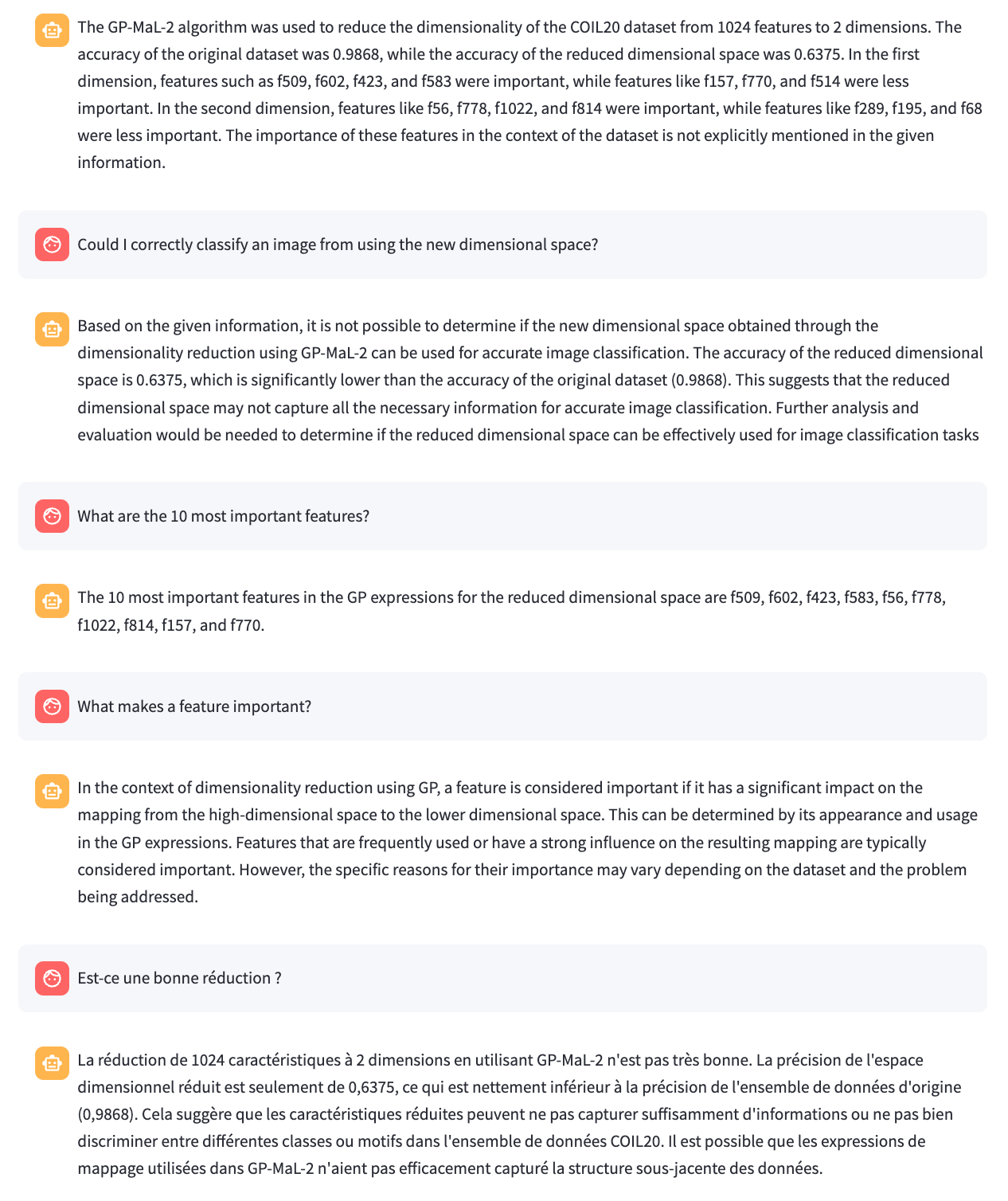}
\caption{COIL20 Case Study Chat Conversation Examples}
\label{CaseStudy3_Chat}
\end{figure*}
\end{document}